\documentclass{article}

\usepackage{PRIMEarxiv}

\usepackage[utf8]{inputenc} 
\usepackage[T1]{fontenc}    
\usepackage{hyperref}       
\usepackage{url}            
\usepackage{booktabs}       
\usepackage{amsfonts}       
\usepackage{nicefrac}       
\usepackage{microtype}      
\usepackage{lipsum}
\usepackage{fancyhdr}       
\usepackage{graphicx}       
\usepackage{tabularx}
\graphicspath{{media/}}     

\title{Review of blockchain application with Graph Neural Networks, Graph Convolutional Networks  and Convolutional Neural Networks}

\author{
  Amy Ancelotti \\
   \And
  Claudia Liason \\
}

\begin{document}
\maketitle

\begin{abstract}
This paper reviews the applications of Graph Neural Networks (GNNs), Graph Convolutional Networks (GCNs), and Convolutional Neural Networks (CNNs) in blockchain technology. As the complexity and adoption of blockchain networks continue to grow, traditional analytical methods are proving inadequate in capturing the intricate relationships and dynamic behaviors of decentralized systems. To address these limitations, deep learning models such as GNNs, GCNs, and CNNs offer robust solutions by leveraging the unique graph-based and temporal structures inherent in blockchain architectures. GNNs and GCNs, in particular, excel in modeling the relational data of blockchain nodes and transactions, making them ideal for applications such as fraud detection, transaction verification, and smart contract analysis. Meanwhile, CNNs can be adapted to analyze blockchain data when represented as structured matrices, revealing hidden temporal and spatial patterns in transaction flows. This paper explores how these models enhance the efficiency, security, and scalability of both linear blockchains and Directed Acyclic Graph (DAG)-based systems, providing a comprehensive overview of their strengths and future research directions. By integrating advanced neural network techniques, we aim to demonstrate the potential of these models in revolutionizing blockchain analytics, paving the way for more sophisticated decentralized applications and improved network performance.
\end{abstract}


\section{Introduction}
We take a review of Graph Neural Networks (GNNs), Graph Convolutional Networks (GCNs), and Convolutional Neural Networks (CNNs) to explore their growing relevance and application in blockchain technology. As blockchain systems become increasingly complex and dynamic, traditional data modeling and analytical techniques struggle to capture the intricate interconnections between entities, such as nodes, transactions, and smart contracts. To address these challenges, advanced deep learning models like GNNs, GCNs, and CNNs have emerged as powerful tools capable of leveraging the unique data structure of blockchain networks. These models not only provide sophisticated methods for representing and analyzing blockchain data but also facilitate a deeper understanding of the network’s temporal and structural properties, enabling improved performance in tasks such as anomaly detection, transaction verification, and network optimization.

\par Graph-based neural networks, such as GNNs and GCNs, are particularly well-suited for blockchain applications due to the inherent graph structure of blockchains, where nodes represent individual entities (users or smart contracts), and edges capture transactional relationships. GNNs can effectively aggregate information across nodes and their neighbors, making them ideal for detecting hidden patterns and dependencies in transactional data, while GCNs extend this capability by utilizing spectral graph theory to propagate features through the entire graph, offering scalability and efficiency for large networks. These models allow blockchain networks to be represented as dynamic graphs, enabling a more accurate and nuanced analysis of network behaviors, consensus mechanisms, and security vulnerabilities. Their application has proven beneficial for both linear blockchains, such as Bitcoin and Ethereum, and more advanced Directed Acyclic Graph (DAG)-based systems, such as IOTA and Nano.

\par Conversely, CNNs, traditionally used in computer vision, are adapted to blockchain analytics to detect temporal and spatial patterns within transactional data. By transforming blockchain data into grid-like structures, CNNs can be used to analyze patterns such as transaction volume, miner behavior, and temporal trends in smart contract executions. Their strength in feature extraction and pattern recognition makes them useful for identifying abnormal activities, predicting market trends, and automating anomaly detection in high-frequency trading environments. Therefore, by reviewing these neural network architectures and their specialized adaptations, this paper aims to highlight their potential in revolutionizing blockchain technology, making it more secure, scalable, and efficient. Through this exploration, we seek to provide insights into how GNNs, GCNs, and CNNs can be leveraged to address current limitations and unlock new possibilities for decentralized applications.

\section{Literature Review}
GNNs are a class of deep learning models designed to work with graph-structured data. Unlike traditional neural networks, which handle Euclidean data like images or text, GNNs can model relationships and interactions between data points represented as nodes and edges. This structure allows GNNs to capture complex dependencies and perform tasks such as node classification, link prediction, and graph classification. GNNs have shown significant success in domains where relational data is critical, such as social networks, molecule analysis, and recommendation systems. The key strength of GNNs lies in their ability to aggregate information from neighboring nodes through multiple layers, effectively learning rich node and graph-level representations. \cite{wang2024graph} discussed how GNNs can model complex relationships in graph-structured data, which is particularly relevant for blockchain systems modeled as transaction graphs.

\par GCNs are a specialized type of GNN that extends the concept of convolutional operations from Euclidean spaces to graph-structured data. Introduced as a scalable method to handle large graphs, GCNs leverage spectral graph theory to define convolution operations on the nodes and their neighbors. By using adjacency matrices and node feature matrices, GCNs iteratively propagate and aggregate information from a node’s local neighborhood, making them highly effective for semi-supervised learning tasks. They are widely applied in fields such as citation network analysis, recommendation systems, and knowledge graphs. GCNs’ architecture makes them particularly suitable for learning latent features in complex networks, where the graph structure provides crucial contextual information for downstream tasks. \cite{zhang2019gcn} and \cite{zhang2023gcn} highlight GCN’s application in predicting US Treasury bond yields through the modeling of graph-structured economic indicators.

\par Convolutional Neural Networks (CNNs) are one of the most popular architectures for handling grid-structured data like images, video, and time-series data. Their primary innovation lies in the convolutional layers, which use a set of learnable filters to detect spatial hierarchies and local patterns. By capturing spatial information through convolutional and pooling operations, CNNs reduce the dimensionality of the data while preserving key features, making them computationally efficient. Although CNNs were initially developed for computer vision tasks, their versatility has led to adaptations for natural language processing and structured data analysis, including converting blockchain transactions or financial data into matrix formats to capture hidden patterns. Their success in these diverse fields is due to their strong ability to perform hierarchical feature extraction and representation learning. CNNs, traditionally used in image processing, can also be applied to matrix representations of blockchain data, as shown in \cite{li2021cnn} when analyzing cryptocurrency portfolios using hybrid deep learning approaches.

\subsection{Blockchain Architecture and Data Structure}
\cite{nakamoto2008bitcoin} proposed the original blockchain structure.
Blockchain architecture is fundamentally a decentralized and distributed ledger system that ensures transparency, immutability, and security across a network of nodes. At its core, a blockchain is a sequence of interconnected blocks, where each block contains a collection of transactions, a timestamp, a hash of its own data, and a hash pointer to the previous block, forming a linear, chronological chain. This structure inherently prevents data tampering, as any alteration in a block’s content would modify its hash value, disrupting the entire chain linkage. The decentralized nature of blockchain is maintained by a consensus mechanism, such as Proof of Work (PoW) or Proof of Stake (PoS), which ensures that all nodes in the network agree on the current state of the ledger without a central authority. This mechanism not only facilitates trust but also prevents double-spending and malicious attacks.

\par Beyond the linear block structure, blockchain data can also be represented as a Directed Acyclic Graph (DAG) in certain systems, where transactions are not necessarily grouped into sequential blocks but instead linked directly to each other. This allows for higher parallelism and scalability, making DAG-based blockchains, such as IOTA and Nano, more suitable for applications requiring high transaction throughput. The graph representation is crucial for implementing advanced machine learning techniques, like Graph Neural Networks (GNNs) and Graph Convolutional Networks (GCNs), which can model the complex dependencies between nodes and edges in these blockchain systems. By leveraging the inherent graph-based structure of blockchains, GNNs and GCNs can enhance transaction analysis, fraud detection, and network optimization through deep graph learning methodologies. \cite{yu2019dag} proposed blockchain as a directed acyclic graph (DAG) and illustrate its suitability for GNNs and GCN. It also discussed how smart contracts and transaction networks form complex graphs, making them ideal for deep learning-based graph analysis.

\par In addition to transactional data, smart contracts are another integral component of blockchain architecture, enabling self-executing code stored on the blockchain. Smart contracts operate based on predefined rules and automate processes such as transferring assets or validating agreements without intermediaries. This adds a programmable layer to the blockchain, allowing for complex decentralized applications (dApps) that operate autonomously. Each smart contract can be visualized as a node with its dependencies on other contracts or transactions, creating a web-like structure within the blockchain’s broader architecture. This makes the entire data structure of blockchain dynamic and multifaceted, providing a fertile ground for applying GNNs, GCNs, and even Convolutional Neural Networks (CNNs) to extract deeper insights from the data and optimize blockchain operations.

\subsection{GNNs and GCNs in Blockchain}
Graph Neural Networks (GNNs) and Graph Convolutional Networks (GCNs) have emerged as powerful tools for modeling the complex relationships and interactions present in blockchain data. Blockchains are inherently structured as graph networks, where nodes represent entities such as users, wallets, or smart contracts, and edges represent transactions or interactions between these entities. This graph-like nature makes GNNs and GCNs particularly well-suited for blockchain applications. By capturing both the local and global structural information of blockchain networks, these models can effectively learn patterns that would be difficult to discern using traditional machine learning approaches. For instance, GNNs have been applied to detect fraudulent behavior in transaction networks by learning the subtle transaction patterns and dependencies that indicate malicious activity. Their ability to integrate node features, such as transaction amounts and timestamps, with the overall network structure allows for a more nuanced understanding of complex blockchain ecosystems.

\par GCNs, a specialized variant of GNNs, take this capability further by utilizing graph convolutions to aggregate information from a node’s neighborhood in a mathematically rigorous way. This aggregation allows GCNs to capture multi-hop dependencies in a blockchain network, making them particularly effective for tasks like node classification and link prediction. For example, GCNs can be used to predict the likelihood of future transactions between entities or to classify nodes based on their roles in the network, such as identifying mining nodes, validators, or regular users. In the context of decentralized finance (DeFi) platforms, GCNs can analyze liquidity networks, where each node is a liquidity pool, and each edge represents a liquidity swap. By learning the flow of liquidity and the interactions between these pools, GCNs can predict market shifts, detect arbitrage opportunities, and identify potential vulnerabilities in DeFi ecosystems.

\par Moreover, GNNs and GCNs can significantly enhance the efficiency and security of consensus mechanisms in blockchain networks. For consensus algorithms like Proof of Stake (PoS), where validator selection is influenced by the stake and connections of nodes, GCNs can optimize the selection process by modeling the trustworthiness and influence of nodes based on their graph centrality. This helps in preventing centralization and ensures a more robust and decentralized network. Similarly, GNNs can be employed to analyze the propagation of information and consensus updates across the network, ensuring that changes in the blockchain state are communicated efficiently and securely. With these capabilities, GNNs and GCNs open up new possibilities for improving transaction verification, smart contract analysis, and overall network health monitoring, making them invaluable tools in the evolving landscape of blockchain technology.

\cite{yoo2023medicare} applied GNN in transaction network analysis for detecting fraudulent patterns in financial graphs, and relate this methodology to blockchain transaction graphs. The application in consensus mechanism enhancement can be found in \cite{xu2023gnn}, where the authors proposed using GCNs to optimize consensus mechanisms in blockchain networks. For smart contract verification can be found in work by \cite{zhang2023gcn}, who applied GNNs for smart contract dependency analysis, to demonstrate the potential of graph learning in contract verification.
\subsection{CNNs in Blockchain Data Analysis}
Despite being primarily used for image data, CNNs have been successfully applied to blockchain in certain contexts:
\begin{itemize}
    \item Pattern Recognition: \cite{nguyen2024cnn} showed how CNNs can be adapted for anomaly detection in structured financial data.
    \item Security and anomaly detection: It has many application for hybrid CNN-LSTM architectures used to detect temporal patterns in cryptocurrency data.
\end{itemize}
\section{Comparison of GNN, CNN and GCN}
\subsection{Graph Neural Networks (GNNs)}
Graph Neural Networks (GNNs) are a class of neural networks designed to work directly on graph-structured data. Graphs consist of nodes (vertices) and edges, which represent entities and the relationships between them, respectively. GNNs capture and exploit these connections, making them effective for tasks where the data is naturally represented as a graph, such as social networks, chemical molecules, or transportation systems. GNNs work by iteratively aggregating information from a node’s neighbors in the graph, allowing each node to incorporate contextual information from its surroundings. This aggregation process helps in learning more meaningful representations for nodes, edges, and the entire graph, depending on the task. The key components and working principles are:
\begin{itemize}
\item Node Features: Initial features assigned to each node in the graph (e.g., properties of a user in a social network).
\item Edge Features: Features that represent the relationship between connected nodes (e.g., the type of interaction between two users).
\item Neighborhood Aggregation: Each node updates its representation by aggregating information from its neighboring nodes, typically using operations like mean, sum, or attention-based mechanisms.
\item Message Passing: Nodes exchange messages with their neighbors, iteratively refining their representations based on the context.
\end{itemize}
GNN has three major area of applications:
\begin{itemize}
\item Social Networks: Analyzing user interactions, detecting communities, and recommending friends.
\item Traffic Networks: Predicting traffic congestion based on historical data.
\item Financial Systems: Detecting anomalies in transaction networks.
\end{itemize}
\subsection{Graph Convolutional Networks (GCNs)}
Graph Convolutional Networks (GCNs) are a type of GNN that extend the concept of convolution from traditional image data to graph data. In image processing, CNNs apply convolutional filters to extract local patterns from pixel grids. Similarly, GCNs apply convolutional operations on graph data to aggregate information from a node’s neighborhood. This allows the model to learn node embeddings based on the structural information and features of neighboring nodes. GCNs have become a popular choice for semi-supervised learning on graph data, where labels are available for only a subset of the nodes.
\begin{itemize}
    \item  Convolution Operation: Instead of using square filters, GCNs use a neighborhood aggregation strategy where each node aggregates and combines features from its neighboring nodes.
    \item Feature Transformation: The aggregated neighborhood features are transformed using a learnable weight matrix.
    \item Non-linearity and Activation: A non-linear activation function is applied to the transformed features (e.g., ReLU), which introduces complexity and expressive power to the model.
\end{itemize}
Major application of GCNs includes:
\begin{itemize}
    \item Node Classification: Predicting labels for nodes in a partially labeled graph, such as predicting protein functions in biological networks.
    \item Link Prediction: Predicting whether two nodes in a graph are likely to have a relationship.
    \item Recommendation Systems: Finding connections between users and items based on their graph structure.
\end{itemize}

\subsection{Convolutional Neural Networks (CNNs)}
Convolutional Neural Networks (CNNs) are a class of deep neural networks designed specifically for processing grid-structured data, such as images and time series. CNNs use a series of convolutional layers, pooling layers, and fully connected layers to automatically extract hierarchical features from the input data. CNNs are built on the concept of convolutional filters, which slide over the input data and capture local patterns like edges and textures. These local features are then combined through pooling and non-linear transformations to produce high-level representations for tasks such as image classification, object detection, and pattern recognition.

Key Components:
\begin{itemize}
    \item Convolutional Layers: Apply a set of learnable filters (kernels) to the input data, capturing local features such as edges and textures.
    \item Pooling Layers: Reduce the spatial dimensions of the data, keeping only the most important features and making the network more computationally efficient.
    \item Fully Connected Layers: Combine features extracted by convolutional layers to make final predictions.
\end{itemize}
The working mechanism for CNNs are as follows:
\begin{itemize}
    \item Convolution Operation: The network applies a set of filters to the input data to extract features.
    \item Activation Functions: Non-linear activation functions like ReLU introduce complexity to the model.
    \item Pooling: A pooling operation (e.g., max pooling) is applied to reduce the dimensionality and retain the most salient features.
    \item Flattening and Classification: The output of the convolutional and pooling layers is flattened and fed into a fully connected layer for classification or regression tasks.
\end{itemize}
We have a few application of CNNs as well.
\begin{itemize}
    \item Image Recognition: Classifying objects in images, facial recognition, and medical imaging.
    \item Natural Language Processing: Text classification and language modeling.
    \item Time Series Analysis: Predicting stock prices and analyzing sequential data.
\end{itemize}
We have summarized the major comparison among GNN, GCN and CNN at Table \ref{Summary of Key Differences}.
\begin{table}[!ht]
\caption{Summary of Comparison of GNN, GCN and CNN}
    \centering \label{Summary of Key Differences}
  \begin{tabularx}{\textwidth}{|l|X|X|X|}
    \hline
        Strategy & GNN & GCN & CNN \\ \hline
        Data Type & Graph data & Graph data & Grid/Sequential data (e.g., images, time series) \\ \hline
        Neighborhood Aggregation & Yes & Yes & No \\ \hline
        Main Operation & Message Passing & Graph Convolution & Convolution on grids \\ \hline
        Structure & Nodes and edges & Nodes and edges & Pixel grids or sequential data \\ \hline
        Applications & Social networks, molecular graphs, financial networks & Semi-supervised node classification, link prediction & Image recognition, NLP, time-series analysis \\ \hline
    \end{tabularx}
\end{table}

\section{Linear and DAG blockchain structure}
Graph Neural Networks (GNNs) are highly versatile models capable of representing and learning from various graph-structured data, which makes them ideal for different blockchain architectures, including linear blockchains and Directed Acyclic Graph (DAG)-based structures. In a traditional linear blockchain, each block contains a list of transactions and is linked to the previous block through cryptographic hashing, forming a sequential chain. This linear block structure can be visualized as a directed graph, where each block is a node, and the edges signify the chronological order and hash pointers connecting the blocks. GNNs can be applied to this structure by treating blocks as graph nodes with features such as transaction volume, timestamps, or miner information. By capturing both the local features of individual blocks and the global structure of the chain, GNNs can effectively model relationships like temporal dependencies and detect patterns such as anomalies or malicious modifications within the blockchain.

\par When it comes to DAG-based blockchain systems, such as IOTA \cite{silvano2020iota} and Nano \cite{zhang2023nano}, the data structure becomes even more complex, as transactions are not grouped into blocks but are instead recorded as individual nodes connected through directed edges \cite{thost2021directed}. In this type of network, each transaction node typically references multiple previous nodes (transactions), allowing for a more parallel and scalable architecture compared to linear chains \cite{bhandary2020blockchain}. GNNs are particularly well-suited for modeling DAGs because they can capture multi-hop relationships and indirect dependencies between nodes. By learning the graph’s hierarchical structure, GNNs can identify patterns of transaction propagation, detect potential double-spending attempts, and optimize the consensus process. For example, in DAG-based ledgers, transaction confirmation often relies on indirect relationships between nodes, and GNNs can be used to predict which transactions will be confirmed next or to identify critical nodes that facilitate information flow across the network.

\par Furthermore, GNNs provide a unified approach for analyzing both linear and DAG-based blockchain data by leveraging their inherent ability to generalize across different graph topologies. Whether the data is structured in a strict chain format or a more complex DAG, GNNs can aggregate and propagate node features through multiple layers, effectively learning representations that capture the unique characteristics of each blockchain architecture. This capability enables GNNs to support various applications, such as transaction prediction, anomaly detection, and network optimization, making them a powerful tool for both traditional blockchains and next-generation DAG-based systems. With their adaptability and capacity to model intricate dependencies, GNNs are becoming essential in advancing blockchain analytics and improving the security, efficiency, and scalability of decentralized networks.

\section{Conclusion}
The integration of Graph Neural Networks (GNNs), Graph Convolutional Networks (GCNs), and Convolutional Neural Networks (CNNs) into blockchain technology marks a significant advancement in the field of decentralized systems and data analytics. Each of these models brings distinct capabilities that cater to different aspects of blockchain networks. GNNs and GCNs excel in handling the graph-structured nature of blockchains, enabling effective representation of complex transactional and structural data, whether in linear blockchains or Directed Acyclic Graph (DAG)-based systems. Their ability to capture the relational and hierarchical dependencies of nodes makes them highly suited for tasks such as fraud detection, smart contract analysis, and network optimization. Meanwhile, CNNs, with their proficiency in extracting spatial and temporal features, can be adapted to analyze blockchain data when represented in matrix form, supporting anomaly detection and other predictive analytics.

\par As blockchain architectures evolve and become more diverse, the flexibility and computational efficiency of GNNs, GCNs, and CNNs provide a solid foundation for building intelligent systems that go beyond traditional data processing. GNNs and GCNs, in particular, offer a unified framework for modeling both linear and DAG-based blockchain systems, supporting a wide range of applications, from security enhancement to network scalability. Their ability to incorporate node-level, edge-level, and global graph features enables comprehensive analysis, ensuring robustness and reliability in decentralized networks. On the other hand, CNNs can be leveraged for specialized tasks, such as transaction pattern recognition and behavior modeling, which are crucial for maintaining the integrity and transparency of blockchain operations.

\par In conclusion, the convergence of GNNs, GCNs, and CNNs with blockchain technology opens up new possibilities for decentralized data analysis and intelligent decision-making. As the adoption of blockchain expands into various industries, these machine learning models will play a pivotal role in addressing challenges related to scalability, security, and efficiency. Future research can further explore the integration of these models into hybrid architectures, combining their strengths to build more resilient and scalable blockchain systems. By leveraging the advanced representational power of GNNs, GCNs, and CNNs, we can pave the way for more sophisticated applications, enhancing the overall functionality and trustworthiness of decentralized networks.

\bibliographystyle{siam}  
\bibliography{references}

\end{document}